# GMOCSO: Multi-objective Cat Swarm Optimization Algorithm based on a Grid System

Aram M. Ahmed[1], Bryar A. Hassan[1,2], Tarik A. Rashid[1], Kaniaw A. Noori[4], Soran Ab. M. Saeed[4], Omed H. Ahmed[5], Shahla U. Umar[6]

[1]Computer Science and Engineering Department, University of Kurdistan Hewlêr, Erbil, Iraq.
[2]Department of Computer Science, College of Science, Charmo University, Chamchamal, Sulaimani,
[4]Database Technology Department, Technical College of Informatics, Sulaimani Polytechnic University, Sulaimani, Iraq.
[5]Department of Information Technology, University of Human Development, Sulaymaniyah, Iraq.
[6]Network Department, College of Computer Science and Information Technology, Kirkuk University, Kirkuk, Iraq.
Corresponding author (Email): Bryar A. Hassan, bryar.ahmad@ukh.edu.krd; bryarahassan@gmail.com

Abstract:
This paper presents a multi-objective version of the Cat Swarm Optimization Algorithm called the Grid-based Multi-objective Cat Swarm Optimization Algorithm (GMOCSO). Convergence and diversity preservation are the two main goals pursued by modern multi-objective algorithms to yield robust results. To achieve these goals, we first replace the roulette wheel method of the original CSO algorithm with a greedy method. Then, two key concepts from Pareto Archived Evolution Strategy Algorithm (PAES) are adopted: the grid system and double archive strategy. Several test functions and a real-world scenario called the Pressure vessel design problem are used to evaluate the proposed algorithm's performance. In the experiment, the proposed algorithm is compared with other well-known algorithms using different metrics such as Reversed Generational Distance, Spacing metric, and Spread metric. The optimization results show the robustness of the proposed algorithm, and the results are further confirmed using statistical methods and graphs. Finally, conclusions and future directions were presented..

## 1. Introduction:

The term optimization means trying to select the best solution for a specific problem among many alternative solutions. The objective can be minimization, such as minimizing cost or time, or it can be maximization, such as maximizing profit or production [1]. In terms of Mathematics, optimization problems can be formulated in a generic form like follows:

$$\min_{x \in R^n} f_i(x), (i = 1,2,\dots,I)$$

$$\text{Subject to} \begin{cases} c_j(x) = 0, (j = 1,2,\dots,J) \\ c_k(x) \geq 0, (k = 1,2,\dots,K) \end{cases} \quad (1)$$

Here if we suppose that the parameters of the search space are denoted in a vector like "$x$", then $f_i(x)$ will be the objective functions or sometimes called fitness functions.

$f(x)$ can be a minimizing or maximizing function depending on the design on hand. Finally, $c_j(x)$ and $c_k(x)$ are the equality and inequality constraints of the design problem respectively [2].

Optimization Problems can be classified according to the objectives, constraints, landscape, form functions, and variables. When the classification is based on the number of objectives, there are two kinds of optimization problems, which are single and multi-objective. If equation (1) is taken, the problem has a single objective, when $I = 1$. This means that there is only one goal in the search space and all the factors are used to reach that goal. However, when $I > 1$, it is a multi-objective problem, which contains multiple conflicting objectives or goals. Many of the real-world problems are of multi-objective types, for example, maximizing profits and meanwhile minimizing time or cost. When these objectives clash with each other, algorithms require having a trade-off between them, which leads to the generation of a set of compromised solutions. This is commonly known as Pareto-optimal solutions [3]. In addition, constraints in optimization problems are those conditions that a solution must fulfill. Optimization problems can also be classified according to the number of these constraints. For example, in the previous equation: if $j = k = 0$, this means there are no constraints and it is known as an "unconstrained optimization problem". As for the equation, if $j > 0$ or $k > 0$, the problem is known as "equality-constrained optimization problem" or "inequality-constrained optimization problem" respectively. Furthermore, depending on the nature of the problems, some models may require a discrete set of values to be used for representing the solution sets [4]. These types of optimization are called discrete optimization. However, some other models, known as continuous optimization, are only comprehended when actual real numbers are employed to represent the solution sets. It is worth mentioning that continuous optimization algorithms can also be used to solve discrete problems. Moreover, problems can be called linear if the change in the output is proportional to the change in the input. On the other hand, nonlinear problems are exactly the opposite, and thus their outputs are unpredictable, and this makes them harder to be solved. Finally, Unimodal problems are those problems that have one global optimum in the search space, while multimodal problems have multiple global optima in the search space [4].

When dealing with multi-objective problems, Decision, and Objective Spaces need to be taken into consideration. The decision space is a space that contains all the possible solutions, while the evaluations of these solutions are presented in a space called objective space. It is worth mentioning that, in real-world scenarios, some of these solutions, which are known as infeasible regions, are not applicable due to the design limitations of the problem [5]. In single objective optimization problems, one can easily compare two solutions and decide which one is better or more optimum. However, in multi-objective optimization problems, the concept of dominance is required to compare the solutions.

Let $x$ and $y$ be two solutions for a minimization multi-objective optimization problem, as shown in equation (2). Thus, the objectives are as follows:

$$\min_{x \in X}(f1(x), f2(x), f3(x), \ldots, fk(x)) \qquad (2)$$
$$\min_{y \in X}(f1(y), f2(y), f3(y), \ldots, fk(y))$$

Where $k \geq 2$ and the set $X$ includes all the feasible solutions.

Then: $x$ dominates $y$ if:

1. $fi(x) \leq fi(y)$ for all indices $i \in \{1, 2, \ldots, k\}$
2. $fj(x) < fj(y)$ for at least one index $j \in \{1, 2, \ldots, k\}$

Therefore, $x$ dominates $y$ i.e. $x \prec y$ if:

1. Solution $y$ is not better than solution $x$ in any of the objectives

2. Solution $x$ is better than solution $y$ in at least one objective.

In the process of multi-objective optimization, solutions are often obtained, where neither of them dominates the others i.e. one solution dominates the other for an objective function, and meanwhile, the other solution dominates this solution for a different objective function. These types of solutions, which cannot be dominated by the other solutions, are known as non-dominated solution set [6].

In addition, the non-dominated solution set belonging to the whole feasible decision space is known as the Pareto front. Those Pareto front solutions that cannot be further improved are called True Pareto front. If an evolutionary multi-objective algorithm is not able to find the True Pareto front, it will attempt to obtain an approximate set of solutions that is close to it. This set of solutions is called the Approximate Pareto Front.

Several methods have been offered for multi-objective problems. Schafer initially proposed the novel notion of using evolutionary optimization methods for multi-objective optimization in 1984 [2]. Ever since many scholars have tried to create various multi-objective algorithms. For example, Non-dominated Sorting Genetic Algorithm 2 (NSGA-II) [7], Multi-objective Particle Swarm Optimization (MOPSO) [8], Multi-objective Evolutionary Algorithm based on Decomposition (MOEA/D) [9], Pareto Archived Evolution Strategy (PAES) [10], and Strength Pareto Evolutionary Algorithm (SPEA/SPEA2) [11] are among the most well-known algorithms in the literature. The cat Swarm Optimization Algorithm is a robust metaheuristic swarm-based algorithm, which is originally proposed by Chu et al. in 2006 [12]. Pradhan and Panda proposed multi-objective Cat Swarm Optimization (MOCSO) by extending CSO to deal with multi-objective problems [13]. MOCSO is combined with the concept of the external archive and Pareto dominance to handle the non-dominated solutions. Orouskhani et al. proposed a Multiobjective algorithm by combining the CSO algorithm with Borda ranking [14]. They also proposed an enhanced version by combining the CSO algorithm with an opposition-based learning technique and the elitism idea [15]. Dwivedia et al. extended the quantum-inspired cat swarm optimization as a Multiobjective algorithm, which uses the idea of Qubits [16]. Murtza et al. extended the integer cat swarm optimization as a multi-objective algorithm, which can be used for binary multi-objective problems [17].

Multi-objective algorithms have two main goals to pursue while searching for optimum solutions; these goals are convergence and diversity preservation [18]. In the first one, the Approximate Pareto front has to move towards the True Pareto front as much as possible, and hence, the algorithm obtains better and more optimum solutions. However, in Diversity preservation, the set of solutions in the Approximate Pareto front has to be well distributed in the area. The word "well distribute" means that they have to cover the whole area of the Pareto front and have uniform distances in between as much as possible. The reason behind this goal is to provide the Decision maker with solutions from all areas and let him/her choose their region of interest (ROI). Therefore, this paper presents a multi-objective version of the Cat Swarm Optimization Algorithm, which is called the Grid-based Multi-objective Cat Swarm Optimization Algorithm (GMOCSO). To accomplish these objectives, we start by substituting a greedy method for the original CSO algorithm's roulette wheel mechanism. The grid structure and double archive method are then incorporated as essential ideas from the Pareto Archived Evolution Strategy Algorithm (PAES). The pressure vessel design challenge is a real-world case study that is utilized along with several test functions to gauge how well the suggested method performs. Using various measures, including Reversed Generational Distance, Spacing metric, and Spread metric, the experiment compares the proposed method against other well-known algorithms such as Mayfly Optimization Algorithm (MMA) [19], Multi-objective Dragonfly Algorithm (MODA) [20] and Strength Pareto Evolutionary Algorithm 2 (SPEA2) [11]. The rest of the paper is organized as follows: Section 2 presents the proposed GMOCSO algorithm. Section 3 discusses the Performance evaluation framework for the GMOCSO algorithm. Section 4 shows the obtained results and finally, section 5 provides the conclusion and future directions.

## 2. Grid-based Multi-objective Cat Swarm Optimization Algorithm (GMOCSO):

The proposed algorithm adopts two key concepts from the PAES algorithm [10] to extend the CSO algorithm into a multi-objective scheme. First, the proposed algorithm has two populations called internal and external populations. The internal population is used to store the initial population and the dominated individuals, whereas the external archive (repository) is used to store the non-dominated individuals in each iteration. This external archive is empty at the beginning and has a predefined size that cannot be exceeded. Second, the proposed algorithm uses a hyper-grid system to determine the least crowded area in the Pareto front of the objective space. This hyper-grid scheme divides the objective space into several non-overlapping hyper-boxes, where each box might contain several agents. Figure (1) explains the hyper-grid system.

The fitness values of these agents denote their coordinates in the objective space. Consequently, this can be used to specify which agent belongs to which hyper-box. Agents inside the least crowded box are called leaders and one of them is used as a global attraction mechanism, where the rest of the agents tend to move towards it. In addition, the roulette wheel method of the CSO algorithm is replaced with the greedy method to drive the agents to move toward the global optimum and hence increase the exploitation ratio of the algorithm.

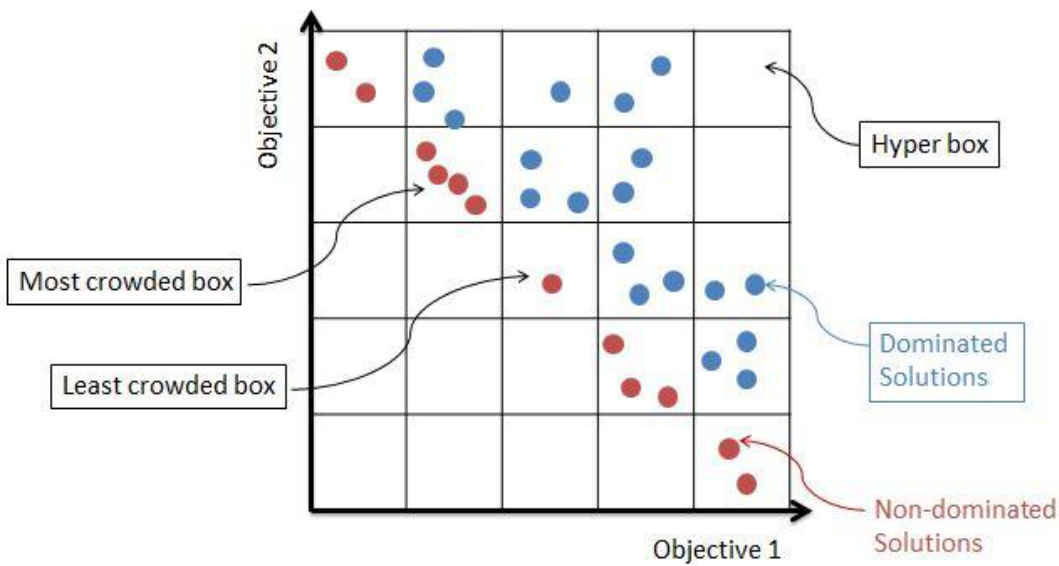

**Figure 1**. Explains the hyper-grid system.

Therefore, the algorithm takes the following steps in the optimization process:

1. Randomly generate the initial population of cats.
2. Calculate the fitness cost for all the cats.
3. Determine the non-dominated cats and store them in an external archive.
4. Use the hyper-grid system to create the hyper boxes.
5. The archive has a predefined size (for example archive=100); If the archive was overfilled, select the most crowded box and delete the extra members.
6. Then, select the least crowded box and randomly select a member from the box to be the leader.
7. Send the leader to the tracing mode.
8. Assign all the cats into the tracing mode.
    - Update velocities for all the cats using equation (3).

$$V_{k,d} = V_{k,d} + c_1 * \text{rand} * (X_{leader,d} - X_{k,d}) \quad (3)$$

Where $V_{k,d}$ is the velocity for the $k^{th}$ cat in the $d^{th}$ dimension; $X_{leader,d}$ is the position of the leader cat, which was chosen from the least crowded box; $X_{k,d}$ is the current position of the $k^{th}$ cat in the $d^{th}$ dimension. $c_1$ is a constant and $rand$ is a single uniformly distributed random number in the range of [0, 1].

   - Update positions for all the cats using equation (4).

$$X_{k,d} = X_{k,d} + V_{k,d} \quad (4)$$

Where $X_{k,d}$ is the position of a cat in the $d^{th}$ dimension.

9. Assign the cats into the seeking mode:
    - Use CDC and SMP parameters to generate the candidate positions, where the original position is one of the candidates.
    - Determinate the non-dominated candidate position to be a new position; if there were more than one non-dominated candidate, randomly select one of them.
10. If any cat dimension is out of the boundary, it is equal to the limits, as in equation (5):

$$X_{k,d} = \begin{cases} L_d & \text{if } X_{k,d} < L_d \\ U_d & \text{if } X_{k,d} > U_d \end{cases} \quad (5)$$

Where $L_d$ and $U_d$ are lower and upper bounds for the search space.

11. Check if the termination condition is satisfied, then terminate the program. Otherwise, repeat Step 2 to Step 10, see Figure (2)

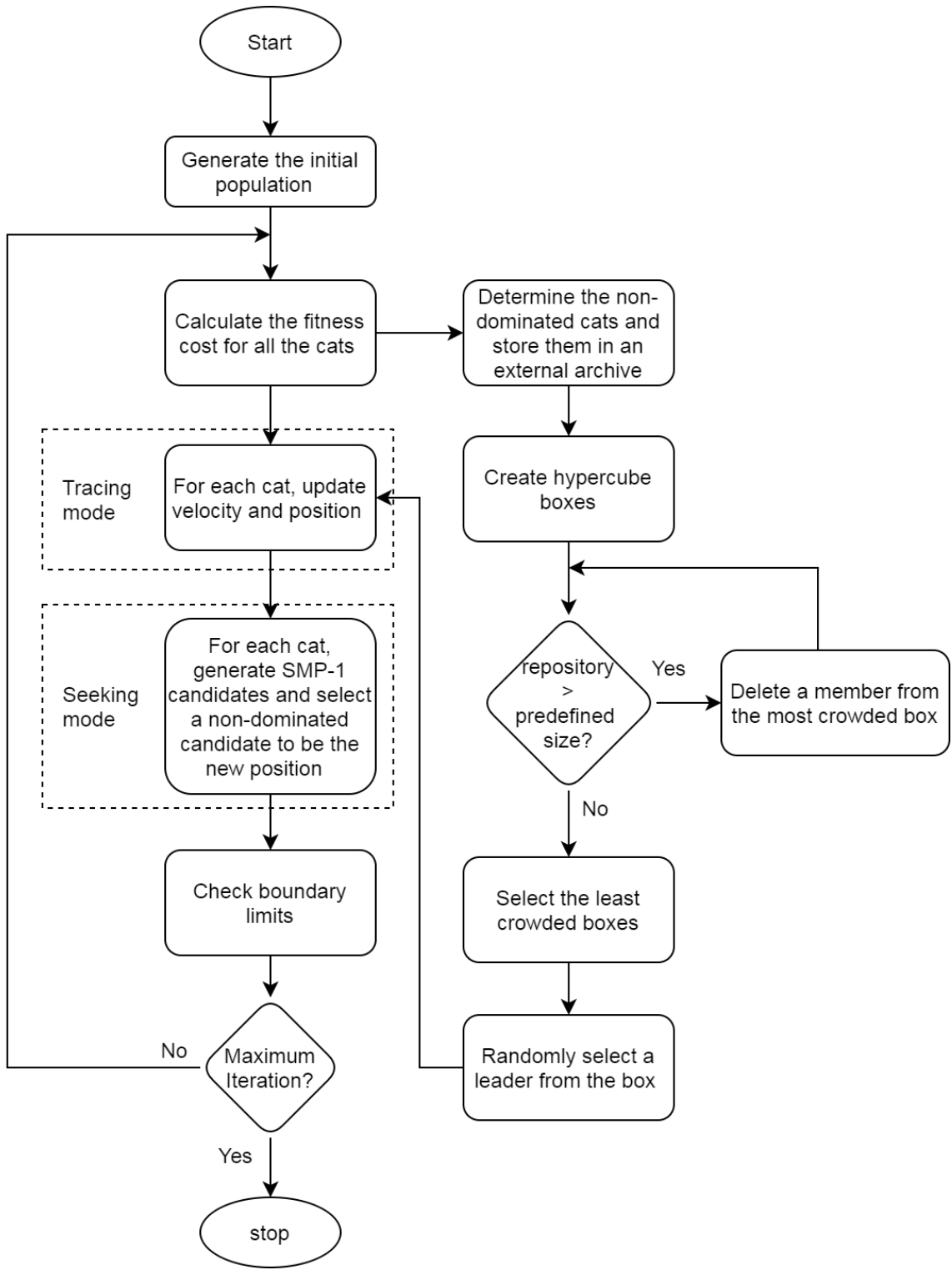

**Figure 2**. Flowchart of the GMOCSO algorithm.

## 3. Performance evaluation framework for the GMOCSO algorithm

In this section, the performance of the proposed GMOCSO algorithm is evaluated using a general performance evaluation framework as shown in Figure (3). The framework consists of two different criteria as follows:

i. Applying the algorithm on ZDT benchmark functions.
ii. Applying the algorithm to a real-world scenario.

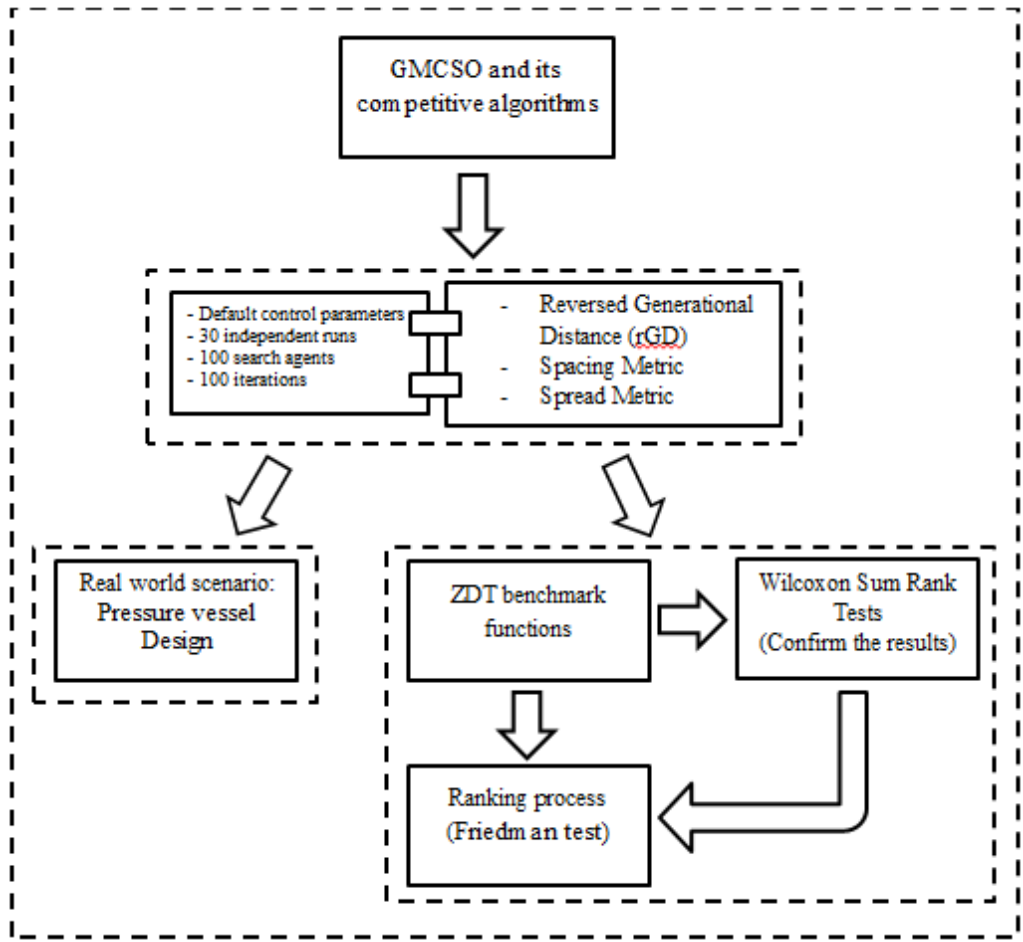

**Figure 3.** The general performance evaluation framework for the GMOCSO algorithm.

### 3.1.1 Benchmark functions

To assess the performance of the proposed GMOCSO algorithm, it was first applied to five common benchmark functions which were ZDT1, ZDT2, ZDT3, ZDT4, and ZDT6 [21]. Table (1) presents their mathematical details

**Table 1**. Mathematical descriptions for the ZDT functions.

| Function name | Mathematical definition |
|---|---|
| ZDT1 | $f_1(x) = x_1$ |
| | $f_2(x) = g(x)\left[1 - \sqrt{x_1/g(x)}\right]$ <br><br> Where $g(x)$ is defined as: <br><br> $g(x) = 1 + 9\left(\sum_{i=2}^{n} x_i\right)/(n-1)$ |
| ZDT2 | $f_1(x) = x_1$ <br><br> $f_2(x) = g(x)\left[1 - \left(\frac{x_1}{g(x)}\right)^2\right]$ <br><br> Where $g(x)$ is defined as: <br><br> $g(x) = 1 + 9\left(\sum_{i=2}^{n} x_i\right)/(n-1)$ |
| ZDT3 | $f_1(x) = x_1$ <br><br> $f_2(x) = g(x)\left[1 - \sqrt{x_1/g(x)} - \frac{x_1}{g(x)}\sin(10\pi x_1)\right]$ <br><br> Where $g(x)$ is defined as: <br><br> $g(x) = 1 + 9\left(\sum_{i=2}^{n} x_i\right)/(n-1)$ |
| ZDT4 | $f_1(x) = x_1$ <br><br> $f_2(x) = g(x)\left[1 - \sqrt{x_1/g(x)}\right]$ <br><br> Where $g(x)$ is defined as: <br><br> $g(x) = 1 + 10(n-1) + \sum_{i=2}^{n}[x_i^2 - 10\cos(4\pi x_i)]$ |
| ZDT6 | $f_1(x) = 1 - e^{(-4x_1)}\sin^6(6\pi x_1)$ <br><br> $f_2(x) = g(x)\left[1 - \left(\frac{f_1(x)}{g(x)}\right)^2\right]$ <br><br> Where $g(x)$ is defined as: <br><br> $g(x) = 1 + 9\left[\frac{\sum_{i=2}^{n} x_i}{(n-1)}\right]^{\frac{1}{4}}$ |

**Experiment setting:**

For each benchmark function, the algorithm was executed for 30 independent runs. For each run, the number of both search agents and iterations was equal to 100. Furthermore, the results were compared with three well-known algorithms, namely Multi-objective Mayfly Optimization

Algorithm (MMA) [19], Multi-objective Dragonfly Algorithm (MODA) [20], and Strength Pareto Evolutionary Algorithm 2 (SPEA2) [11]. Parameter settings for these algorithms are presented in Table (2).

**Table 2**. Parameter settings details for the selected Multi-objective algorithms

| ALGORITHMS | PARAMETERS | VALUES |
|---|---|---|
| GMOCSO | POPULATION SIZE | 100 |
| | MAX ITERATION | 100 |
| | INERTIA WEIGHT W(I) | 1 |
| | SMP | 2 |
| | CDC | 1 |
| | CONSTANT (C) | 1 |
| | NO. OF GRID | 10 |
| | SRD | 1 |
| | ARCHIVE SIZE | 100 |
| MMA | POPULATION SIZE | 100 |
| | MAX ITERATION | 100 |
| | INERTIA WEIGHT | 0.8 |
| | INERTIA WEIGHT DAMPING RATIO | 1 |
| | PERSONAL LEARNING COEFFICIENT | 1.0 |
| | GLOBAL LEARNING COEFFICIENT | 1.5 |
| | BETA | 2 |
| | MUTATION COEFFICIENT | DANCE=0.77, |
| | RANDOM FLIGHT | FL=0.77, FL_DAMP=0.99 |
| | NO. OF CROSSOVER | 20 |
| | MUTATION RATE | 0.02 |
| | ARCHIVE SIZE | 100 |
| MODA | POPULATION SIZE | 100 |
| | MAX ITERATION | 100 |
| | INERTIA WEIGHT W(I) | [0.2 – 0.9] |
| | RANDOM VALUES | [0,1] |
| | ARCHIVE SIZE | 100 |
| SPEA2 | POPULATION SIZE | 100 |
| | MAX ITERATION | 100 |
| | ARCHIVE SIZE | 100 |

Then, three essential metrics were used to assess the achieved results, which are listed below:

1. **Reversed Generational Distance**

Reversed Generational Distance or rGD. This metric is calculated by determining the Euclidian distance between the points of the true Pareto front and the closes point on the approximate Pareto distance [22]. The main purpose of this criterion is to show the ability of the different algorithms to determine a set of non-dominated individuals having the shortest distance with the true Pareto optimal fronts. Therefore, it can be seen that algorithms with the smallest generation distance

(rGD) produce the best convergence to the true Pareto front i.e. the approximate Pareto front is the closest to the true Pareto front. Furthermore, the best rGD value that can be achieved by an algorithm is zero, where the approximate Pareto front and the true Pareto front are similar. Equation (6) presents the mathematical details of this metric:

$$rGD(t) = \frac{\sum_{i=1}^{|\mathcal{P}^*(t)|} d_i}{|\mathcal{P}^*(t)|} \quad (6)$$

where di is the Euclidean distance between the i-th member of P*(t) and its closest point from Q(t); P*(t) is the true Pareto front and Q(t) is the obtained Pareto Front.

2. **Spacing metric**

The main purpose of the Spacing metric (S) is to demonstrate how well the non-dominated individuals are distributed along the approximate Pareto front. The best distribution is where the individuals are equally distributed and the distance between them is equal. Equation (7) presents the mathematical details of the Space metric [23].

$$S = \sqrt{\frac{1}{n_{pf}} \sum_{i=1}^{n_{pf}} (d_i - \bar{d})^2} \quad (7)$$

Where npf is the number of members in the approximate Pareto front and di is the Euclidean distance between the member within the approximate Pareto front and the nearest member in the approximate Pareto front. Therefore, the algorithm with the minimum S value has the best distribution of its individuals. In the best case, the S value is equal to zero, where the solutions are completely uniform and the distance between the non-dominated solutions is equal.

3. **Spread metric or Delta metric**

This metric is also called the Delta metric, which is used to show how far the non-dominated solutions have extended and how well they are spread along the approximate Pareto front. Equation (8) presents the mathematical details of this metric:

$$\Delta = \frac{d_f + d_l + \sum_{i=1}^{n_{pf}} |d_i - \bar{d}|}{d_f + d_l + (n_{pf} - 1)\bar{d}} \quad (8)$$

where df and dl are the Euclidean distances between the extreme solutions in the true Pareto front and the approximate Pareto front, respectively. Further, di is the Euclidean distance between each point in the approximate Pareto front and the closest point in the true Pareto front. npf and d¯ are defined as the total number of members in the approximate Pareto front and the average of all distances, respectively [23].

Subsequently, based on the results that can be achieved from the three metrics, the Friedman test was used to rank the algorithms and determine their overall performances. For this, the average values that were achieved from the 30 independent runs were used. Meanwhile, the Wilcoxon sum rank test is used to calculate P-values and confirm the result.

Apart from these well-known metrics, another metric called elapsed time measurement was also used. The elapsed time calculates the amount of time each algorithm requires to finish one independent run.

### 3.1.2 Appling the GMOCSO algorithm on a real-world scenario

The proposed GMOCSO algorithm was also applied to a real-world scenario called the Pressure vessel design problem [24]. A cylindrical pressure vessel is topped at both ends by curved heads as presented in Figure (4).

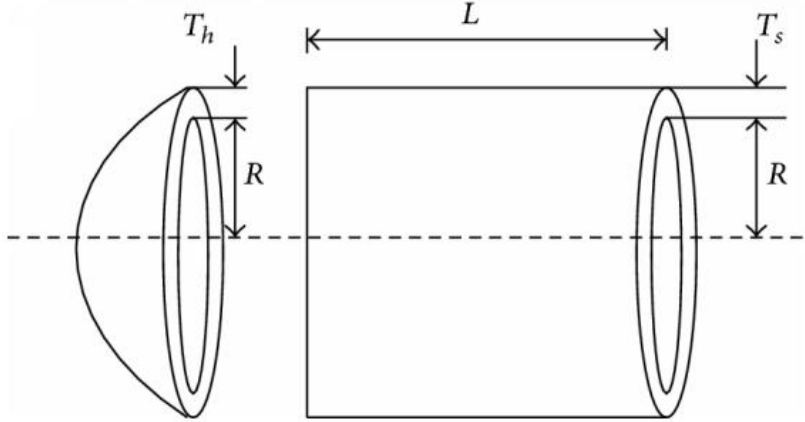

**Figure 4**. Pressure vessel design problem, adopted from [24]

The problem considers the design of an air storage tank with a working pressure of 1000 psi and a minimum volume of 750 ft3.

The first objective of the problem is to reduce the total cost of a cylindrical pressure vessel. This cost involves the cost of material and the cost of forming and welding) :

$$f_1(x) = 0.6224x_1x_3x_4 + 1.7781x_2x_3^2 + 3.1661x_1^2x_4 + 19.84x_1^2x_3, \qquad (9)$$

Subject to:

$$g_1(x) = x_1 - 0.0193x_3 \geq 0,$$

$$g_2(x) = x_2 - 0.00954x_3 \geq 0,$$

$$g_3(x) = \pi x_3^2 x_4 + \frac{4}{3}\pi x_3^3 - 1296000 \geq 0,$$

Where $x_1, x_2 \in \{1, \ldots, 100\}$, $x_3 \in [10,200]$, and $x_4 \in [10,240]$. $x_1$ and $x_2$ are integer multiples of 0.0625.

$x_1 = T_s$ , which represents the thickness of the shell

$x_2 = T_h$ , which represents the head of the pressure vessel

$x_3 = R$ , which represents the inner radius

$x_4 = L$, which represents the length of the cylindrical section

The second objective of the problem is the sum of the three constraint violations:

$$f_2(x) = \sum_{i=1}^{3} max\{g_i(x), 0\} \qquad (10)$$

The experiment settings and the parameter settings are all the same as in the previous section.

## 4. Results and Discussions:

Based on the evaluation criteria mentioned in the previous section, the following results are obtained, which are presented below:

### 4.1 Results for benchmark functions:

The proposed GMOCSO algorithm was first benchmarked on six test functions called the ZDT functions. In the experiment, three more algorithms were chosen for comparison, namely the multi-objective Mayfly optimization algorithm (MMA) [19], the multi-objective Dragonfly Algorithm (MODA) [20], and Strength Pareto Evolutionary Algorithm 2 (SPEA2) [21].

The results of the benchmark functions are presented in Table (3).

Table 3. Comparison results between GMOCSO and the Selected Algorithms for the ZDT Test Functions in terms of Average and Standard Deviation.

| F | Metrics | GMOCSO | | MMA | | MODA | | SPEA2 | |
|---|---|---|---|---|---|---|---|---|---|
| | | Average | STD | Average | STD | Average | STD | average | STD |
| ZDT1 | RGD | 0.008251645 | 0.000645212 | **0.004779893** | 0.000222114 | 0.100799368 | 0.09082163 | 0.053992 | 0.017841 |
| | S | 0.077806532 | 0.011078946 | 0.075503344 | 0.006606084 | **0.027136347** | 0.044158279 | 0.878917 | 0.238826 |
| | Delta | 0.853871219 | 0.032721397 | **0.399518442** | 0.039293129 | 1.364857245 | 0.160130313 | 0.853213 | 0.099897 |
| ZDT2 | RGD | 0.028268889 | 0.109917947 | **0.004856085** | 2.07E-04 | 0.074477018 | 0.062369162 | 34254.15 | 8481.733 |
| | S | 0.075916272 | 0.016613567 | 0.076431697 | 0.006518955 | **0.017571498** | 0.013397005 | 463969.7 | 94282.35 |
| | Delta | 0.839039408 | 0.052525346 | **0.395613038** | 0.029195821 | 1.428002616 | 0.127234485 | 0.827969 | 0.139153 |
| ZDT3 | RGD | 0.009046667 | 0.001527072 | **0.005300991** | 0.000211342 | 0.086483057 | 0.062537721 | 0.116075 | 0.027901 |
| | S | 0.217135677 | 0.025091761 | 0.269998037 | 0.02158768 | **0.080144325** | 0.084676525 | 0.895159 | 0.260512 |
| | Delta | 0.939763609 | 0.080065515 | **0.559479212** | 0.016761967 | 1.345371429 | 0.127584933 | 0.784542 | 0.100351 |
| ZDT4 | RGD | 8.22E-03 | 0.000785028 | **0.004776174** | 0.000244893 | 1.83E-01 | 0.096584815 | 5.01E-02 | 0.014521 |
| | S | **8.22E-03** | 0.000785028 | 0.075192952 | 0.006955307 | 2.11E-01 | 0.491719343 | 6.30E+00 | 1.252434 |
| | Delta | 8.62E-01 | 0.040205206 | **0.39837872** | 0.033276381 | 1.28E+00 | 0.186181632 | 9.18E-01 | 0.174281 |
| ZDT6 | RGD | 0.007520491 | 0.001834684 | **0.003779973** | 0.000120089 | 0.030610889 | 0.031934169 | 0.134059 | 0.08455 |
| | S | 0.256361178 | 0.310232383 | 0.233597523 | 0.339582389 | **0.158387045** | 0.258385158 | 3.093395 | 0.51416 |
| | Delta | 1.016366296 | 0.201172776 | **0.589973634** | 0.382808442 | 1.518800475 | 0.158049857 | 0.809812 | 0.144761 |

Based on the results achieved in Table (3), the Freidman test was conducted to rank the algorithms and determine the overall performance of the algorithms. According to the results of the test, which are presented in Table (4) and Figure (5), GMOCSO ranks second. Then, the Wilcoxon sum rank test was also conducted to find the P-values and further confirm the results. The results of the Wilcoxon sum rank test are shown in Table (5). As can be seen, the P-values for all the results are less than 0.05, except for four cases which are highlighted in bold. So, it can be concluded that the results are significant.

Finally, the Algorithms were compared using the elapsed time measurement. The results are presented in Table (6). As it is shown, the proposed GMOCSO algorithm is the most efficient and takes the least amount of time, while the MMA algorithm takes the most amount of time. Therefore, it can be concluded that even though the GMOCSO algorithm performs worse than the MMA algorithm in terms of the RGD, S, and Delta metrics, it recompenses the loss by its efficiency.

**Table 4**. Ranking of the selected algorithms for the ZDT test functions (Friedman test).

| TF | Metrics | Ranking GMOCSO | Ranking MMA | Ranking MODA | Ranking SPEA2 |
|---|---|---|---|---|---|
| ZDT1 | RGD | 2 | 1 | 4 | 3 |
| | S | 3 | 2 | 1 | 4 |
| | Delta | 3 | 1 | 4 | 2 |
| ZDT2 | RGD | 2 | 1 | 3 | 4 |
| | S | 2 | 3 | 1 | 4 |
| | Delta | 3 | 1 | 4 | 2 |
| ZDT3 | RGD | 2 | 1 | 3 | 4 |
| | S | 2 | 3 | 1 | 4 |
| | Delta | 3 | 1 | 4 | 2 |
| ZDT4 | RGD | 2 | 1 | 4 | 3 |
| | S | 1 | 2 | 3 | 4 |
| | Delta | 2 | 1 | 4 | 3 |
| ZDT6 | RGD | 2 | 1 | 3 | 4 |
| | S | 3 | 2 | 1 | 4 |
| | Delta | 3 | 1 | 4 | 2 |
| RGD subtotal | | 10 | 5 | 17 | 18 |
| RGD ranking | | 2 | **1** | 3.4 | 3.6 |
| S subtotal | | 11 | 12 | 7 | 20 |
| S ranking | | 2.2 | 2.4 | **1.4** | 4 |
| Delta subtotal | | 14 | 5 | 20 | 11 |
| Delta ranking | | 2.8 | **1** | 4 | 2.2 |
| Total | | 35 | 22 | 44 | 49 |
| Overall ranking | | 2.333333 | **1.466667** | 2.933333 | 3.266667 |

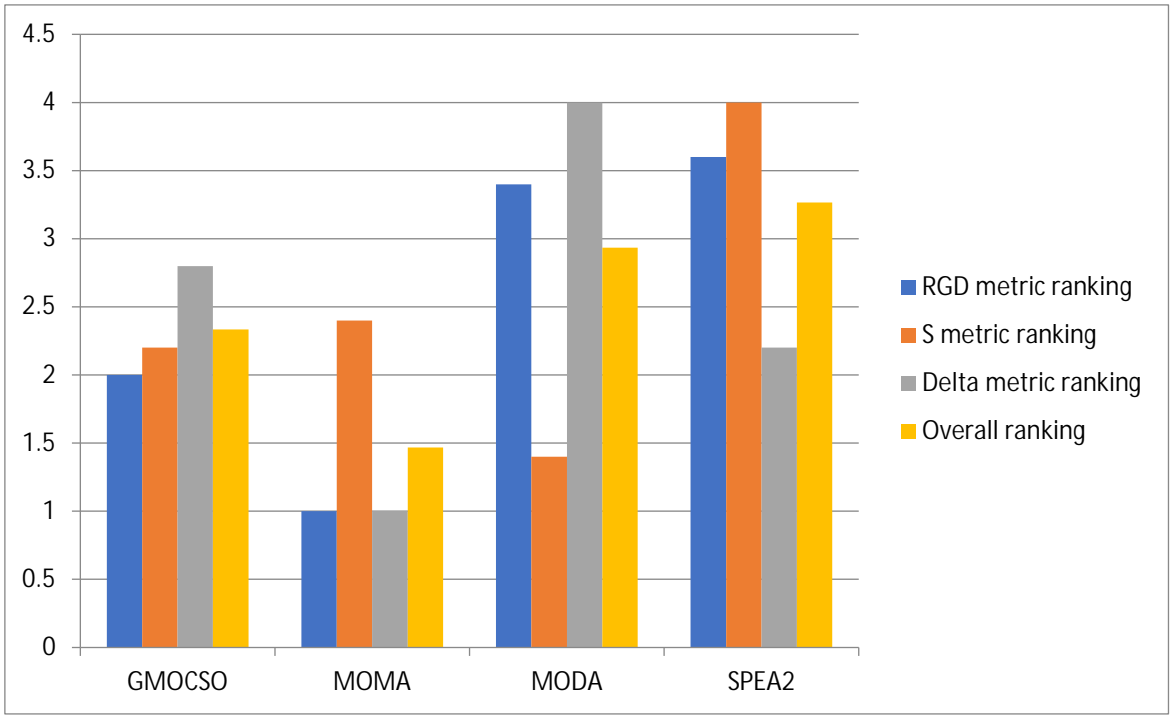

**Figure 5**. Ranking of the selected algorithms in the experiment for the ZDT test functions using the Friedman test.

**Table 5**. Wilcoxon Sum Rank Tests between the GMOCSO algorithm and the selected algorithms for the ZDT test functions.

| Functions | Metric | GMOCSO vs. MMA (P-values) | GMOCSO vs. MODA (P-values) | GMOCSO vs. SPEA2 (P-values) |
|---|---|---|---|---|
| ZDT1 | RGD | 1.691123e-17 | 1.691123e-17 | 1.691123e-17 |
| | S | **0.708193** | 5.998246e-13 | 1.691123e-17 |
| | Delta | 1.691123e-17 | 1.691123e-17 | **0.820070** |
| ZDT2 | RGD | 1.691123e-17 | 4.841517e-13 | 4.841517e-13 |
| | S | **0.320778** | 4.841517e-13 | 3.017967e-11 |
| | Delta | 1.691123e-17 | 1.691123e-17 | **0.431747** |
| ZDT3 | RGD | 1.691123e-17 | 1.691123e-17 | 1.691123e-17 |
|  | S | 5.798703e-11 | 4.035171e-09 | 1.691123e-17 |
|  | Delta | 1.691123e-17 | 2.029348e-16 | 1.727306e-08 |
| ZDT4 | RGD | 1.691123e-17 | 1.691123e-17 | 1.691123e-17 |
|  | S | 1.691123e-17 | 1.085811e-10 | 1.691123e-17 |
|  | Delta | 1.691123e-17 | 1.120285e-12 | 0.035767 |
| ZDT6 | RGD | 1.691123e-17 | 7.610055e-16 | 1.691123e-17 |
|  | S | 0.004530 | 0.000452 | 1.691123e-17 |
|  | Delta | 0.000011 | 9.745944e-14 | 0.000054 |

**Table 6**. Comparison results between the GMOCSO and the Selected Algorithms for the ZDT test function in terms of elapsed time measurement.

| Functions | Elapsed time for GMOCSO (seconds) | Elapsed time for MMA (seconds) | Elapsed time for MODA (seconds) | Elapsed time for SPEA2 (seconds) |
|---|---|---|---|---|
| ZDT1 | **23.18582** | 141.5477 | 28.87808 | 100.2301 |
| ZDT2 | **24.16883** | 140.7262 | 29.17383 | 101.101 |
| ZDT3 | **22.78487** | 143.8483 | 23.47593 | 100.472 |
| ZDT4 | **22.40294** | 141.8431 | 29.39307 | 100.5177 |
| ZDT6 | **19.91391** | 150.2155 | 27.35512 | 100.095 |

### 3.1.1 Results and Discussions for the Pressure Vessel Design Problem

The GMOCSO algorithm was also applied to a real-world scenario called the Pressure Vessel design problem. Similar to the previous section, the proposed algorithm was compared with three more algorithms, which are the multi-objective Mayfly optimization algorithm (MMA) [19], multi-objective Dragonfly Algorithm (MODA) [20], and Strength Pareto Evolutionary Algorithm

2 (SPEA2) [21]. The algorithms were run 30 independent times, and each time, the number of search agents and iterations was 100. Then, the average value and standard deviation of these 30 independent runs were taken. Furthermore, the algorithms were also compared using the elapsed time measurement. The results are all presented in Table (7). It can be noticed, that the proposed GMOCSO algorithm yields very competitive results.

**Table 7**. Comparison results between the GMOCSO and the Selected Algorithms for the Pressure Vessel Design problem.

| Algorithms | Metrics | Pressure vessel design problem. | | | |
| --- | --- | --- | --- | --- | --- |
| | | RGD | S | Delta | Elapsed time (Second) |
| GMOCSO | Average | **12450.82** | 337249.75 | 0.912582 | 22.47755 |
| | STD | 13194.535 | 58915.063 | 0.168256 | |
| MMA | Average | 20813.508 | 309937.61 | 1.0772376 | 116.2052 |
| | STD | 13987.115 | 52101.146 | 0.131381 | |
| MODA | Average | 19041.145 | **114826.8** | 1.5380991 | 24.88461 |
| | STD | 4497.0885 | 51727.737 | 0.0962503 | |
| SPEA2 | Average | 34254.15 | 463969.7 | **0.827969** | 100.3218 |
| | STD | 8481.733 | 94282.35 | 0.139153 | |

## 5. Conclusion:

This paper presents a multi-objective version of the Cat Swarm Optimization Algorithm, which is called the Grid-based Multi-objective Cat Swarm Optimization Algorithm (GMOCSO). For this, two main modifications were made. First, the seeking mode of the algorithm was modified and the roulette wheel approach was replaced with a greedy method. This technique increases the convergence rate of the algorithm. Second, to preserve the diversity of the non-dominated solutions in the objective space, two key concepts from the PAES algorithm were adopted, which were the external population and the hyper-grid box techniques. In addition, several benchmark functions and a real-world scenario called the Pressure Vessel Design problem were used to evaluate the performance of the proposed algorithm. The results were then compared with three other well-known algorithms. Next, the Friedman test and the Wilcoxon sum rank were used to rank the algorithms and confirm the results. Furthermore, the elapsed time measurement was used to test the efficiency of the algorithm. Finally, it can be concluded that the proposed algorithm can efficiently converge towards the Pareto front solutions and still preserve its diversity along the area. In the future, one can replace the hyper grid box with the indicator-based method to obtain better diversity perseverance. We have noticed that in recent years, optimization algorithms are still considered one of the most promising techniques in terms of being integrated with other technologies. This means that the changes and developments in it are continuing to come up with new methods that can be more effective and capable of yielding more satisfying results. For future reading, the authors advise the reader could optionally read the following research works wing research works [25]-[42].

**Acknowledgments:**

The authors would like to thank the University of Technology, Baghdad and University of Kurdistan-Hewler for providing facilities for this research work.

**Funding:** This study was not funded.

**Compliance with Ethical Standards Conflict of Interest:** The authors declare that they have no conflict of interest.

**Ethical Approval:** This article does not contain any studies with human participants or animals performed by any of the authors.

**Data Availability Statement:** The datasets generated during and/or analysed during the current study are not publicly available but are available from the corresponding author on reasonable request